\pdfoutput=1

\documentclass[11pt]{article}
\usepackage[preprint]{acl}
\usepackage{enumitem}
\usepackage{times}
\usepackage{latexsym}
\usepackage{indentfirst}
\usepackage{xcolor}
\usepackage{graphicx}
\usepackage[T1]{fontenc}
\usepackage{titlesec}
\titlespacing*{\section}{0pt}{0.4\baselineskip}{0.4\baselineskip}

\usepackage[utf8]{inputenc}

\usepackage{microtype}

\usepackage{inconsolata}

\usepackage{graphicx}
\title{Sample-Efficient Language Model for Hinglish Conversational AI}

\author{\normalfont
  \begin{tabular}{c c c}
    Sakshi Singh & Abhinav Prakash & Aakriti Shah \\
    \normalfont sakshisi@usc.edu & \normalfont ap42546@usc.edu & \normalfont shahaakr@usc.edu \\
    \multicolumn{3}{c}{} \\
    \multicolumn{3}{c}{Chaitanya Sachdeva \quad Sanjana Dumpala} \\
    \multicolumn{3}{c}{\normalfont csachdev@usc.edu \quad dumpala@usc.edu} \\
  \end{tabular}
}

\begin{document}
\maketitle
\begin{abstract}
This paper presents our process for developing a sample-efficient language model for a conversational Hinglish chatbot. Hinglish, a code-mixed language that combines Hindi and English, presents a unique computational challenge due to inconsistent spelling, lack of standardization, and limited quality of conversational data. This work evaluates multiple pre-trained cross-lingual language models, including Gemma3-4B and Qwen2.5-7B, and employs fine-tuning techniques to improve performance on Hinglish conversational tasks. The proposed approach integrates synthetically generated dialogues with insights from existing Hinglish datasets to address data scarcity. Experimental results demonstrate that models with fewer parameters, when appropriately fine-tuned on high-quality code-mixed data, can achieve competitive performance for Hinglish conversation generation while maintaining computational efficiency.
\end{abstract}

\section{Introduction}
Hinglish, a code-mixed hybrid of Hindi and English, has emerged as a prevalent form of communication among bilingual speakers in India and regions with significant Indian populations. Despite its widespread usage in day-to-day conversations, social media, and entertainment, computational resources for Hinglish remain scarce compared to monolingual counterparts. Code-mixed languages present a unique challenge for natural language processing (NLP) tasks, including inconsistent spelling and grammar conventions, as well as frequent language switching patterns.

This work is motivated by the increasing use of Hinglish in digital communication, the lack of conversational AI systems equipped to handle code-mixed dialogue, and the need for resource-efficient models over large, computation-heavy architectures. In addition, there is a growing demand for culturally appropriate agents that reflect the natural communication patterns of bilingual users.

This paper develops a sample-efficient language model for Hinglish conversation to advance the capabilities of NLP systems for code-mixed languages while maintaining computational feasibility. This work addresses the technical challenges of processing non-standardized linguistic inputs and generating contextually appropriate, natural-sounding code-mixed responses.

\section{Related Work}

Recent advancements in low-resource neural machine translation demonstrate promising approaches for languages with limited data availability. \cite{matzopoulos2025babylms} Matzopoulos et al. (2025) showed that small parameter language model (LM) architectures (ELC-BERT and MLSM) achieved significant performance gains in isiXhosa, a low resource language, with only 13M words of training data. Their implementation of ELC-BERT  delivered a +3.2 F1 improvement in NER tasks while requiring 70\% less training time than alternative models, establishing a small-parameter LM as a viable foundation for resource-efficient NLP in morphologically complex languages.

In parallel, \cite{fjoshi2024adapting} Raviraj et al. developed Nemotron-Mini-Hindi 4B, a bilingual small language model focusing on pairs of Hindi-English language. Through continued pre-training and synthetic data augmentation on 400B tokens, their model achieved state-of-the-art results on Hindi benchmarks while maintaining strong English performance, demonstrating that targeted pre-training enhances factual accuracy and language understanding in low-resource contexts. 

This work differentiates itself by applying a small-parameter LM—a significantly more compact and computationally efficient architecture—to the development of a Hinglish( Hindi-English code-mixed language) chatbot.

\section{Problem Description}
This work aims to develop a computationally efficient language model that generates natural, contextually appropriate responses in Hinglish, forming the foundation of a Hinglish chatbot. Our work addresses key challenges in this space.

First, data scarcity is a major issue: most existing Hinglish datasets consist of isolated utterances, lacking the multi-turn structure necessary to model coherent dialogue and context retention. Furthermore, Hinglish’s inconsistent spelling, romanization, and code-switching patterns introduce significant noise.

Resource limitations further complicate model development, as large models require substantial computation and data. This motivates the exploration of smaller architectures that can still perform effectively. Finally, standard NLP metrics often fail to reflect the quality of code-mixed text, which requires us to depart from conventional evaluation methods and explore alternative approaches.

This study investigates how to fine-tune smaller models for Hinglish generation, how to compensate for limited conversational data, and how to better evaluate Hinglish dialogue quality.

\section{Methods}
\subsection{Materials}
During the initial planning phase, we evaluated several publicly available datasets to build a Hinglish conversational corpus. These included \cite{srivastava_singh_hinge} \textit{HinGE} (Hinglish-English parallel data), \cite{lince_dataset} \textit{LinCE} (code-switching evaluation), and the \cite{iiith_hinglish} \textit{IIITH} dataset (conversational code-mixed content). We also considered \cite{hinglishTOP2024} \textit{Hinglish-TOP}, and \cite{phinc2020} \textit{PHINC}, but found them less suitable due to their focus on classification, translation, or limited-turn exchanges. These datasets lacked sufficient conversational depth and did not sufficiently capture the natural dynamics of Hinglish code-switching in spontaneous communication.

Although these resources helped shape our understanding of Hinglish linguistic patterns and annotation practices, none were ultimately used in model training. Instead, we generated our full dataset synthetically, as detailed in the following \textit{Data Procedure} section.

\subsection{Data Procedure}
A culturally grounded Hinglish conversational dataset was constructed entirely through synthetically generated dialogues, using no pre-existing corpora. All training data was generated via API-driven prompting with the \textit{Gemini-2.0-Flash} language model \cite{google_gemini}.

To standardize input, we applied selective normalization on high-frequency romanized Hindi variants (e.g., mapping “bahut”, “bhot”, “bahout” to “bahut”), improving consistency while preserving natural code-mixing patterns. The cleaning steps removed noise such as excessive punctuation, emojis, and artifacts, ensuring cleaner inputs for model training.

We designed a custom Python framework that generated structured, multi-turn dialogues across 40+ everyday topics like \textit{exam stress}, \textit{love}, \textit{roommate} and \textit{many others}. The prompts instructed the LLM to output alternating user-assistant messages in JSON, each turn spanning 40–50 words. Batching, API key rotation, retry logic, and backoff strategies enabled efficient large-scale generation.

This synthetic generation approach allowed us to control linguistic style and conversation quality; producing a dataset tailored to code-mixed Hinglish dialogue. While synthetic data generation provides these positives, ensuring the realism and authenticity of the generated content posed challenges. More than 3,000 dialogues were created (around 5 million tokens). Each was validated for structure, linguistic realism, and topical relevance. Invalid or malformed outputs triggered automated re-prompts until acceptable responses were obtained.

\section{Design}
Multiple pre-trained cross-lingual models were evaluated to find the most suitable architectures for our Hinglish conversational task: Gemma-4B \cite{gemma3_4b_it_2025}, Qwen2.5-7B \cite{qwen2.5-7b-2024}, Qwen2.5-3B \cite{qwen2.5}, Gemma3-1B \cite{gemma3_report_2025}, mT5 \cite{mt5small2021}, GPT-2 \cite{gpt22020}, DistilBERT \cite{distilbert2020}.

Based on our preliminary evaluations, we finalized two primary models for deployment: \textbf{Qwen2.5-3B} and \textbf{Qwen2.5-7B}. Qwen2.5-3B offers an efficient balance for lightweight applications, while Qwen2.5-7B—thanks to its larger capacity—demonstrated a robust understanding of Hinglish inputs and maintained coherence across multi-turn dialogue, making it well suited for code-mixed conversational scenarios. Additionally, \textbf{Gemma3-4B} provides a strong trade-off between model capacity and computational efficiency, and shows promising performance on multilingual generation.

\subsection{Fine-Tuning Approach}
We adopted parameter-efficient fine-tuning techniques like LoRA and QLoRA to reduce memory usage while preserving model quality, enabling training under constrained computational resources. Hyperparameter tuning was performed to optimize learning rates, batch sizes, and training durations. Our strategy also emphasized maintaining contextual coherence across multi-turn Hinglish conversations through targeted fine-tuning objectives.

\section{Experimental Results}

\subsection {Experimental Setup}
Given the lack of quality Hinglish conversational datasets, we generated our own synthetic corpus using controlled prompts to capture natural code-switching. We split this data 80:10:10 for training, validation, and testing.

Our baseline models included Qwen2.5-3B and Qwen2.5-7B, and we also evaluated other models such as Gemma 3-4B. We applied fine-tuning using both LoRA and QLoRA techniques on our custom synthetic Hinglish dataset to explore performance improvements and trade-offs across model sizes.

We intentionally avoided traditional evaluation metrics like BLEU and ROUGE, which are not well-suited for code-mixed languages and penalize valid linguistic variations. Instead, we used a custom evaluation framework focused on code-mixing: Code-Mixing Index (CMI) for language blend, BERTScore for semantic similarity, and cosine similarity tailored to Hinglish.

Human evaluation was central to our assessment. Fluent bilingual evaluators rated model outputs on Hinglish fluency, persona adherence, coherence, language balance, and repetition. This human-first approach was critical to assess the nuances of code-mixed generation that automated metrics often miss.

\subsection{Results and Discussion}
The experimental results demonstrate the effectiveness of parameter-efficient fine-tuning for Hinglish conversational AI. Table~\ref{citation-guide} compares the Qwen2.5 base models with their LoRA-fine-tuned counterparts across key linguistic and cultural metrics.

\textbf{Hinglish Fluency} captures how naturally the model blends Hindi and English, especially in casual, idiomatic code-switching like “haan yaar” or “oye sun.” This is a crucial aspect for conversational models targeting bilingual speakers. The Qwen2.5–3B\_Hinglish\_LoRA\_LD model showed the strongest gains, with a \textbf{41.4\% improvement} in fluency over its base version. The 3B\_Hinglish\_LoRA also showed a \textbf{34.6\% gain}, while the 7B\_Hinglish\_LoRA improved by \textbf{21.9\%}. These results indicate that our fine-tuning approach helps the models better internalize informal, context-aware language blending.

\begin{table}[htbp]
\centering
\renewcommand{\arraystretch}{1.1}
\setlength{\tabcolsep}{7pt} 
\begin{tabular}{|l|c|c|}
\hline
\textbf{Metric (Average)} & \textbf{Base} & \textbf{LoRA} \\
\hline
Hinglish Fluency     & 2.90 & 4.10 (+41.4\%) \\
\hline
Persona Adherence    & 3.20 & 4.10 (+28.1\%) \\
\hline
Gender Correctness   & 4.50 & 4.90 (+8.9\%) \\
\hline
Hindi Usage          & 3.00 & 3.60 (+20.0\%) \\
\hline
Coherence            & 3.30 & 4.70 (+42.4\%) \\
\hline
\end{tabular}
\caption{\label{citation-guide} Qwen2.5 \textbf{3B} Model Comparison}
\end{table}

\vspace{-5pt}

\begin{table}[htbp]
\centering
\renewcommand{\arraystretch}{1.1} 
\setlength{\tabcolsep}{7pt}
\begin{tabular}{|l|c|c|}
\hline
\textbf{Metric (Average)} & \textbf{Base} & \textbf{LoRA} \\
\hline
Hinglish Fluency& 3.20 & 3.90 (+21.9\%) \\
\hline
Persona Adherence    & 3.70 & 3.80 (+2.7\%)  \\
\hline
Gender Correctness   & 4.60 & 5.00 (+8.7\%)  \\
\hline
Hindi Usage          & 3.10 & 3.20 (+3.2\%)  \\
\hline
Coherence            & 4.30 & 4.60 (+7.0\%)  \\
\hline
\end{tabular}
\caption{Qwen2.5 \textbf{7B} Model Comparison}
\end{table}

\textbf{Gender Correctness} reflects how well the model maintains consistent gender references – a critical factor in respectful and accurate communication. Errors in this area can break immersion or be culturally insensitive. The 3B\_Hinglish\_LoRA showed a substantial \textbf{25.6\% improvement}, while the 7B\_Hinglish\_LoRA achieved an \textbf{8.7\% gain}. These improvements suggest that fine-tuning helped the models better understand and retain speaker attributes across turns.

\textbf{Coherence} measures whether the model maintains logical flow and relevance, particularly between multiple conversational exchanges. This is essential to make responses feel natural and context-aware. The 3B\_Hinglish\_LoRA\_LD model led the way here as well, improving coherence by \textbf{42.4\%}. The 3B\_Hinglish\_LoRA improved by \textbf{11.8\%}, while the 7B\_Hinglish\_LoRA saw a \textbf{7.0\% increase}. These gains show that even lightweight fine-tuning significantly enhances the model's ability to stay on-topic and respond smoothly in code-mixed settings.

\begin{figure}[htbp]
\centering
\includegraphics[width=0.47\textwidth]{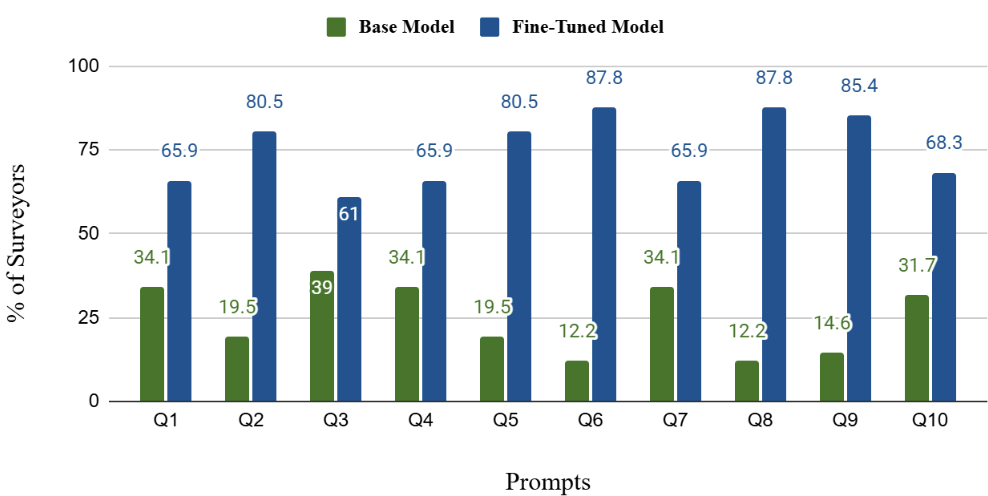}
\caption{\label{citation-guide1} Surveyors' Model Preference}
\label{fig:citation-guide1}
\end{figure}

To further validate these improvements, we conducted \textbf{A/B testing} in which human participants were asked to choose the response that most closely resembled how they personally converse in Hinglish. Across 10 diverse prompts, the fine-tuned model was consistently preferred, with \textbf{surveyor preference rates} as high as \textbf{87.8\%}, compared to just 12-39\% for the base model. This overwhelming preference reflects strong alignment between the fine-tuned model's output and natural Hinglish conversational norms. The results can be seen in Figure~\ref{citation-guide1}.

The generated output was also analyzed quantitatively. The \textbf{average response length} for the fine-tuned model was \textbf{11.3 words}, demonstrating the model's ability to generate concise responses and be efficient in informal settings. Its \textbf{Code-Mixing Index (CMI)} was \textbf{0.692}, indicating a balanced blend of Hindi and English within responses. Furthermore, a \textbf{BERT-F1 score of 0.828} demonstrated a strong semantic similarity to human-generated references, confirming that the fine-tuned outputs remained contextually accurate and relevant.

To benchmark against a state-of-the-art large language model, we also evaluated 100 test prompts using \textbf{LLaMA 70B} and had \textbf{GPT-4o act as the judge}. LLaMA 70B achieved an average score of \textbf{4.6} across all evaluated dimensions (fluency, coherence, correctness, etc.). Our \textbf{Qwen2.5–7B fine-tuned model} scored \textbf{4.1}, while the \textbf{Qwen2.5–3B fine-tuned model} received \textbf{3.86}. These results are notable considering the vast difference in model size: our 7B model approaches the performance of a 70B model, yet is far more efficient in terms of compute and deployability.

Taken together, the qualitative preferences and quantitative metrics reinforce that our parameter-efficient fine-tuning pipeline not only improves technical performance in fluency, coherence, and correctness, but also makes the model’s outputs feel more natural and relatable to real Hinglish speakers. An example conversation and the user interface of the Hinglish chatbot can be seen in Figure~\ref{fig:model_comparison} in the Appendix. These findings underscore the potential of targeted fine-tuning to bring high-quality conversational AI to low-resource, code-mixed language environments without heavy computational costs.

\section{Conclusions and Future Work}
This research demonstrates that high-quality Hinglish conversational chatbots can be built without massive language models or large datasets. By strategically fine-tuning smaller models such as Qwen2.5-7B using parameter-efficient methods like LoRA, we achieved competitive performance with greater computational efficiency.

We found that data quality outweighed quantity, carefully curated and normalized Hinglish data outperformed larger, noisier datasets. Synthetic data generation and additional pre-training on Hinglish text further enhanced the model’s grasp of code-mixed patterns. Overall, our chatbot produced responses comparable to those from larger models.

Some existing evaluation metrics lack nuance for code-mixed languages, and the model sometimes falters on highly colloquial or regional Hinglish. Models with fewer than 3B parameters still showed limited performance, even with quality data.

Future work will explore multimodal and speech-based inputs, interactive fine-tuning via user feedback, and extending our approach to other code-mixed Indian languages like Tanglish (Tamil-English), Benglish (Bengali-English), and Manglish (Malayalam-English). These dialects, like Hinglish, are prevalent in informal communication across India. We also see potential in domain-specific applications like mental health chatbots, content moderation, and e-governance.

Finally, we aim to improve long-form conversational coherence and develop more accurate, code-mixing-aware evaluation frameworks. This work lays the foundation for culturally inclusive, lightweight conversational AI systems that reflect the linguistic realities of multilingual users.

\nocite{*} 
\bibliography{custom}

\begin{thebibliography}{15}
\providecommand{\natexlab}[1]{#1}

\bibitem[{Biswal(2020)}]{phinc2020}
Mrutyunjay Biswal. 2020.
\newblock \href {https://www.kaggle.com/datasets/mrutyunjaybiswal/phincparallel-hinglish-corpus-machine-translation} {Phinc parallel hinglish corpus - machine translation}.
\newblock Kaggle.
\newblock Accessed: 2025-03-03.

\bibitem[{DeepMind(2025{\natexlab{a}})}]{gemma3_4b_it_2025}
Google DeepMind. 2025{\natexlab{a}}.
\newblock Gemma 3 4b instruction-tuned model.
\newblock \url{https://huggingface.co/google/gemma-3-4b-it}.
\newblock Accessed: 2025-04-24.

\bibitem[{DeepMind(2025{\natexlab{b}})}]{gemma3_report_2025}
Google DeepMind. 2025{\natexlab{b}}.
\newblock Gemma 3 technical report.
\newblock \url{https://goo.gle/Gemma3Report}.
\newblock Accessed: 2025-04-24.

\bibitem[{et. al(2024)}]{qwen2.5}
An~Yang et. al. 2024.
\newblock \href {https://arxiv.org/abs/2412.15115} {Qwen2.5 technical report}.
\newblock \emph{arXiv preprint arXiv:2412.15115}.

\bibitem[{Face(2020{\natexlab{a}})}]{distilbert2020}
Hugging Face. 2020{\natexlab{a}}.
\newblock Distilbert.
\newblock \url{https://huggingface.co/docs/transformers/en/model_doc/distilbert}.
\newblock Accessed: 2025-03-05.

\bibitem[{Face(2020{\natexlab{b}})}]{gpt22020}
Hugging Face. 2020{\natexlab{b}}.
\newblock Gpt-2.
\newblock \url{https://huggingface.co/docs/transformers/en/model_doc/gpt2}.
\newblock Accessed: 2025-04-24.

\bibitem[{Google(2024)}]{google_gemini}
Google. 2024.
\newblock \href {https://gemini.google.com/app} {Google gemini app}.
\newblock Accessed: 2025-04-23.

\bibitem[{International Institute~of Information~Technology(2021)}]{iiith_hinglish}
Hyderabad (IIIT-H) International Institute~of Information~Technology. 2021.
\newblock \href {https://www.iiit.ac.in/hinglish/} {Hinglish code-mixed dataset}.
\newblock Accessed: 2025-04-23.

\bibitem[{Matzopoulos et~al.(2025)Matzopoulos, Hendriks, and Marais}]{matzopoulos2025babylms}
Alexis Matzopoulos, Charl Hendriks, and Liezl Marais. 2025.
\newblock {BabyLMs for isiXhosa: Data-Efficient Language Models}.
\newblock \emph{arXiv preprint arXiv:2501.03855}.

\bibitem[{Raviraj et~al.(2024)Raviraj, Kanishk, Anusha, Raunak, Rakesh, Utkarsh, Sanjay, Niranjan, and Eileen}]{fjoshi2024adapting}
Raviraj, Kanishk, Anusha, Raunak, Rakesh, Utkarsh, Sanjay, Niranjan, and Eileen. 2024.
\newblock {Adapting Multilingual LLMs to Low-Resource Languages using Continued Pre-training and Synthetic Corpus}.
\newblock \emph{arXiv preprint arXiv:2410.14815}.

\bibitem[{Research(2021)}]{mt5small2021}
Google Research. 2021.
\newblock mt5-small.
\newblock \url{https://huggingface.co/google/mt5-small}.
\newblock Accessed: 2025-03-05.

\bibitem[{Research(2024)}]{hinglishTOP2024}
Google Research. 2024.
\newblock Hinglish-top dataset.
\newblock \url{https://github.com/google-research-datasets/Hinglish-TOP-Dataset}.
\newblock Accessed: 2025-03-05.

\bibitem[{Srivastava and Singh(2021)}]{srivastava_singh_hinge}
Vivek Srivastava and Mayank Singh. 2021.
\newblock \href {https://aclanthology.org/2021.eval4nlp-1.20/} {Hinge: A dataset for generation and evaluation of code-mixed hinglish text}.
\newblock In \emph{Proceedings of the 2nd Workshop on Evaluation and Comparison of NLP Systems (Eval4NLP)}, pages 200--208, Punta Cana, Dominican Republic. Association for Computational Linguistics.
\newblock Accessed: 2025-04-23.

\bibitem[{Team(2024)}]{qwen2.5-7b-2024}
Qwen Team. 2024.
\newblock Qwen2.5-7b.
\newblock \url{https://huggingface.co/Qwen/Qwen2.5-7B}.
\newblock Accessed: 2025-04-24.

\bibitem[{TheDevastator(2021)}]{lince_dataset}
TheDevastator. 2021.
\newblock \href {https://www.kaggle.com/datasets/thedevastator/unlock-universal-language-with-the-lince-dataset} {Unlock universal language with the lince dataset}.
\newblock Accessed: 2025-04-02.

\end{thebibliography}
\vspace{3\baselineskip}
\appendix

\section{Appendix}
\label{sec:appendix}

This is a screenshot of an example conversation with our Hinglish Chat Buddy.
\begin{figure}[htbp]
\centering
\includegraphics[width=0.47\textwidth]{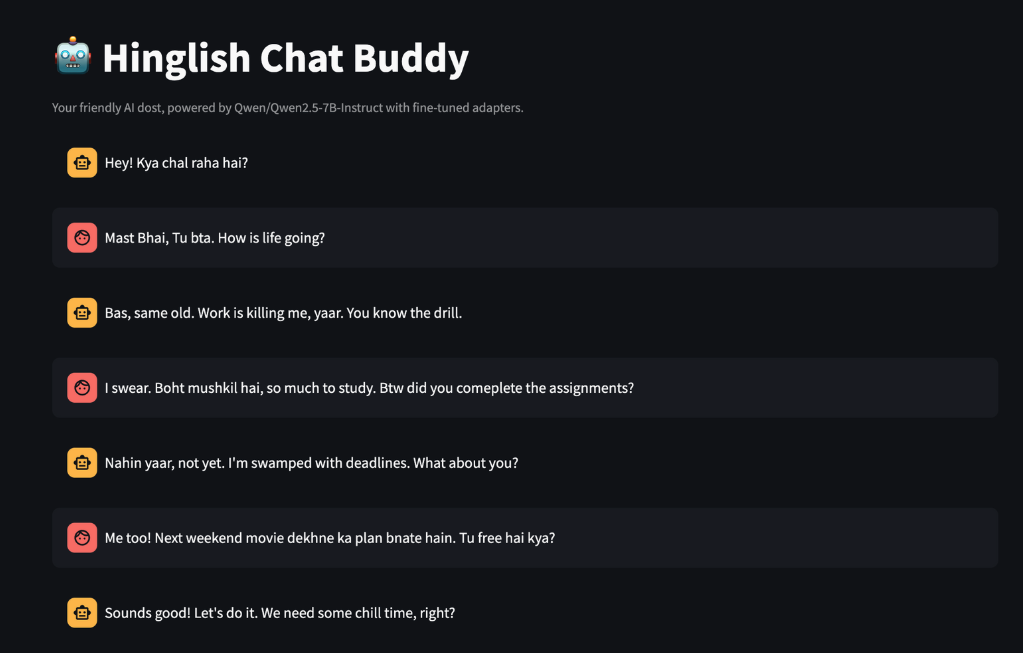}
\caption{\label{model-comparison}Hinglish Chatbot Example Conversation \& UI}
\label{fig:model_comparison}
\end{figure}

\end{document}